\documentclass{article}
\usepackage{graphicx}
\usepackage{amsmath,amssymb}
\usepackage[width=122mm,left=12mm,paperwidth=146mm,height=193mm,top=12mm,paperheight=217mm]{geometry}

\usepackage{subfig}
\usepackage{url}

\begin{document}

\title{Quantifying the amount of visual information used by neural caption generators\footnote{This publication will appear in the Proceedings of the First Workshop on Shortcomings in Vision and Language (2018). DOI to be inserted later.}}

\author{
	Marc Tanti \and Albert Gatt \and Kenneth P. Camilleri\\
	University of Malta, Msida MSD 2080, Malta\\
	\{marc.tanti.06, albert.gatt, kenneth.camilleri\}@um.edu.mt
}

\date{}

\maketitle

\begin{abstract}
This paper addresses the sensitivity of neural image caption generators to their visual input. A sensitivity analysis and omission analysis based on image foils is reported, showing that the extent to which image captioning architectures retain and are sensitive to visual information varies depending on the type of word being generated and the position in the caption as a whole. We motivate this work in the context of broader goals in the field to achieve more explainability in AI.
\end{abstract}

\section{Introduction}
The goal of explainable AI is to move beyond an exclusive focus on the outputs of neural networks, an approach which risks treating such networks as `black boxes' which, though reasonably well-understood at the level of their macro-architecture, are hard to explain because of their complex, non-linear structure \cite{Samek2018}. 

The broad goal of the present paper is to seek a better explanation of the behaviour of neural image caption generators \cite{Bernardi2016}. Such generators typically consist of a neural language model that is conditioned on the features extracted from an image using a convolutional neural network, with several possibilities available on how to do the conditioning \cite{Tanti2018}. 

The main question we address is how sensitive such generators actually are to the visual input, that is, to what extent the string generated by these models varies as a function of variation in the visual features extracted by the image. We address this using a sensitivity analysis \cite{Samek2018} and an analysis based on foils \cite{Shekhar2017}. In addressing this question, we hope to achieve a better understanding of the extent to which caption generation architectures succeed in grounding linguistic symbols in visual features\footnote{Code is available on \url{https://github.com/mtanti/quantifing-visual-information}.}.

\section{Background}

It is known that not all words in a sentence are given equal importance by a neural language model of the kind image captioning systems use \cite{Kadar2017}. 
Rather than measuring the importance of words, as was done in \cite{Kadar2017}, we would like to measure how important the image is at conditioning the language model to emit the different words of a caption during the generation process. This can shed light on the extent to which the generator is grounded in visual data and help to explain some of the model's output decisions.



One way of making neural architectures more explainable is to examine their sensitivity with respect to their inputs \cite{Samek2018}.
Such sensitivity analysis can be done by measuring the gradient of the output with respect to different parts of the input. In this paper, we conduct such an analysis, measuring the gradient of the output with respect to the input image feature vectors.

A related approach is to compare the outcomes produced by a network when parts of the input are altered, or replaced by foils. This has been done in the image captioning domain, and has yielded datasets such as FOIL-COCO \cite{Shekhar2017}. In \cite{Shekhar2017}, the visual sensitivity of images was tested by replacing words in captions with foils and checking if models are able to detect whether a caption contains an incorrect word. The results showed that this is a hard task for many vision-language models, despite being trivial for humans. However, this task does not directly quantify 
the visual sensitivity of such models with respect to different parts of a caption.
This is what we attempt to do in the second part of our analysis.

\section{Data and methods}
\begin{figure}
	\centering
	\subfloat[
		\label{fig:architecture_init}
		Init-inject
	]{
		\includegraphics[scale=0.5]{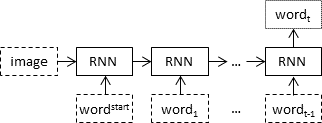}
	}
	\qquad
	\subfloat[
		\label{fig:architecture_pre}
		Pre-inject
	]{
		\includegraphics[scale=0.5]{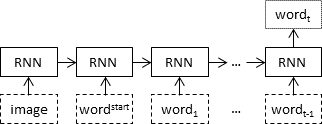}
	}
	
	\subfloat[
		\label{fig:architecture_par}
		Par-inject
	]{
		\includegraphics[scale=0.5]{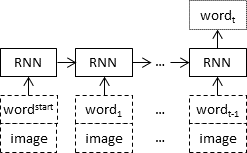}
	}
	\qquad
	\subfloat[
		\label{fig:architecture_merge}
		Merge
	]{
		\includegraphics[scale=0.5]{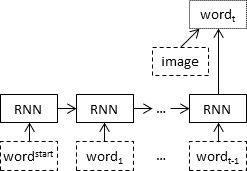}
	}
	\caption{
		\label{fig:architectures}
		\small Different neural image captioning architectures.
	}
\end{figure}

Our goal is to measure how much visual information is used in order to predict a particular word in a caption generator. To do this we make use of the data and models from \cite{Tanti2018}\footnote{See: \url{https://github.com/mtanti/where-image2}}, which examines four different neural caption generator architectures which are often found in the literature. These are illustrated in Fig. \ref{fig:architectures}. They differ mainly in the way the language model is conditioned on image features, as follows:

\begin{itemize}
    \item {\em init-inject}: the image features are used as an initial hidden state vector of the RNN;
    \item {\em pre-inject}: the image features are used as the first input to the RNN;
    \item {\em par-inject}: the image features are included together with every word input to the RNN; and
    \item {\em merge}: the image features are concatenated with the RNN hidden state vector and fed to the softmax layer.
\end{itemize}    


In our experiments, each architecture uses a GRU as an RNN. The architectures were trained on MSCOCO \cite{Lin2014} which was obtained from the distribution provided by \cite{Karpathy2015}\footnote{See: \url{http://cs.stanford.edu/people/karpathy/deepimagesent/}}. The distributed datasets come with the images already converted into feature vectors using the penultimate layer of the VGG-16 CNN \cite{Simonyan2014}. The vocabulary consists of all words that occur at least five times in the training set. We run two sets of experiments to see how much visual information is retained by each architecture: sensitivity analysis and omission scoring; both of which are explained in detail below.

\subsection{Sensitivity analysis}

Sensitivity analysis involves measuring the gradient of a model's output with respect to its input in order to see how sensitive the current output is to different parts of the input \cite{Samek2018}. The more sensitive, the more important that part of the input is to produce the output. We use this technique to measure how sensitive the softmax layer of the caption generator is to the image at different time steps in the generation process. We do this by computing the partial derivative of the softmax output with respect to the input image vector. It is important to note that even though the image might only be input once as an initial state to the RNN, its influence on the output will not be the same at every time step.

As we implemented our neural networks in Tensorflow, which does not currently allow for computing full Jaccard matrices efficiently, instead of finding the gradient of the whole softmax we only take the gradient of the 
probability of the next word in the caption. Measuring the gradient of this word allows us to infer what contribution the image made to the selection of this word during generation. Although the gradient is a single number, it is computed with respect to every element in the image feature vector. We aggregate these partial gradients by taking the mean of the absolute values.

We take captions that were already generated by the same caption generator being analyzed. Each caption is fed back to the caption generator that generated them to re-predict the probability of the next word for every prefix of increasing length in the caption. We report the mean gradient for each time step aggregated over all corresponding time steps in captions of the same length.

We also compare these gradients to the gradient with respect to the last word in the prefix 
(i.e. with respect to the preceding word, but not the image) in order to 
compare a model's sensitivity to linguistic context, as compared to visual features.

\subsection{Omission scoring}
Omission scoring \cite{Kadar2017} 
measures changes in the model's output or hidden layers as some part of an input is removed.
The more similar the hidden layer representation, the less important the removed input is. We use a similar technique to measure how important the image is to the representation. Of course, the image cannot be omitted from a caption generator, but it can be replaced by a 
different image, known as a foil. 
In image caption generators, excluding the merge architecture, the RNN hidden state vector at each time-step 
consists of a mixture of visual information and the preceding caption prefix.
In the case of merge architectures, the same 
mixture
is found in the layer that concatenates the image vector to the RNN hidden state vector. We call these mixed image-prefix vectors `multimodal vectors'.

We take the multimodal vector of a caption generator and measure by how much it changes when a caption prefix is input together with a distractor (foil) image, as opposed to when the correct image is used with that same prefix. This is done after each time step in order to measure whether the distractor image affects the representation less and less over time.

We take captions that were already generated by the caption generator. Each caption is fed back to the caption generator that generated it to re-predict the probability of the next word at every time-step.
This is repeated with a distractor image in place of the correct one.
We then compute the cosine
distance between the multimodal vectors resulting from the correct and the distractor images.
We report the mean cosine distance for each time step aggregated over all corresponding time steps in captions of the same length.

In addition to multimodal vectors, we also compare the softmax layers at each time step with the test image and the foil, to assess the impact of the image change on the output probabilities. Comparison of the softmax layer is done using both cosine distance and Jensen-Shannon divergence.

To identify distractor images, we compare each image in the test set to the others, finding the one whose feature vector is furthest (in terms of cosine) from the correct one.

\section{Results}
The lengths of generated captions vary between 6 and 15 words. For the sake of brevity, we only report the results on captions of length 9, which is the most common length. The results follow the same pattern for captions of other lengths as well.
Table~\ref{tbl:tags} shows the distribution of parts of speech at each of the 9 word positions in the captions; this sheds light on which words cause spikes in visual information usage.

\begin{table}[h]
	\centering
	\begin{small}
		{
		\setlength{\tabcolsep}{0.25em}
		\begin{tabular}{c|cccccccccc}
    		word	& ADJ	& ADP	& ADV	& CONJ	& DET	& NOUN	& NUM	& PRON	& PRT	& VERB \\
    		\hline
    		0	& 	& 	& 	& 	& \textbf{99.8\%}	& 	& 0.4\%	& 	& 	&  \\
    		1	& 22.8\%	& 	& 0.2\%	& 	& 	& \textbf{77.1\%}	& 	& 	& 	& 0.1\% \\
    		2	& 1.5\%	& \textbf{34.1\%}	& 0.1\%	& 7.3\%	& 0.7\%	& 26.2\%	& 	& 	& 	& 30.4\% \\
    		3	& 7.5\%	& \textbf{30.6\%}	& 0.1\%	& 0.5\%	& 9.9\%	& 27.5\%	& 0.1\%	& 0.1\%	& 1.3\%	& 22.6\% \\
    		4	& 3.8\%	& 13.6\%	& 0.1\%	& 1.3\%	& \textbf{33.6\%}	& 23.0\%	& 0.1\%	& 0.2\%	& 1.1\%	& 23.3\% \\
    		5	& 11.8\%	& 16.1\%	& 0.1\%	& 0.5\%	& 6.7\%	& \textbf{52.0\%}	& 0.1\%	& 	& 2.2\%	& 10.5\% \\
    		6	& 0.5\%	& \textbf{51.4\%}	& 	& 4.5\%	& 17.0\%	& 9.3\%	& 0.1\%	& 0.1\%	& 13.7\%	& 3.5\% \\
    		7	& 5.7\%	& 7.1\%	& 	& 1.0\%	& \textbf{68.8\%}	& 13.3\%	& 	& 2.4\%	& 0.1\%	& 1.7\% \\
    		8	& 8.5\%	& 0.1\%	& 0.1\%	& 	& 	& \textbf{88.7\%}	& 	& 2.5\%	& 	& 0.2\% \\
    	\end{tabular}
    	}
	\end{small}
	\caption{
		\small Part of speech tags found at different positions in all 9 word captions. Since different architectures generate different captions, the percentages are averaged over all the architectures. Maximum probability per word position is in bold.
	}
\label{tbl:tags}     
\end{table}

\begin{figure}
	\centering
	
	\subfloat[
		\label{fig:results_sensitivity_image}
		Sensitivity of the next word's probability with respect to the image.
	]{
		\includegraphics[scale=0.6]{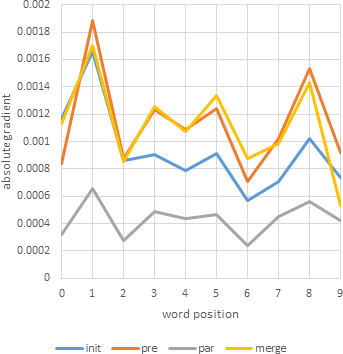}
	}
	\quad
	\subfloat[
		\label{fig:results_sensitivity_prevword}
		Sensitivity of the next word's probability with respect to the previous word.
	]{
		\includegraphics[scale=0.6]{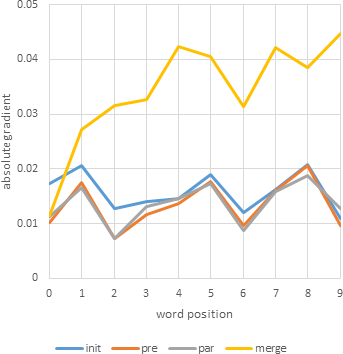}
	}
	\caption{
		\label{fig:results_sensitivity}
		\small Sensitivity analysis of 9-word captions (plus END token). Note that the previous word for position 0 is the START token and position 9 is the END token.
	}
\end{figure}

The results for the sensitivity analysis are shown in Fig.~\ref{fig:results_sensitivity}.
It is clear that certain word positions are more sensitive to the image than others. Irrespective of architecture, there are substantial peaks in Fig.~\ref{fig:results_sensitivity_image} at positions 1 and 8, both of which are predominantly nouns. 
It could be argued that the gradient can be used to detect visual words in captions, that is, nouns referring to objects in the pictures. Par-inject has a consistently low gradient throughout the caption, which is probably reflecting the tendency of the network to avoid retaining excessive visual information, since the same image is input at every time step.

Turning to Fig.~\ref{fig:results_sensitivity_prevword}, the output is much more sensitive to the previous word than to the image, by an order of magnitude. The merge architecture has an upward trend in sensitivity to the previous word as the caption prefix gets longer whilst the other architectures are somewhat more stable. This could be because in the merge architecture, which does not mix visual features directly in the RNN, there is more memory allocated in the RNN to focus on the previous word.

\begin{figure}
	\centering
	\subfloat[
		\label{fig:results_convergence_multimodal_min}
		Omission scores: multimodal vector (cosine distance).
	]{
		\includegraphics[scale=0.5]{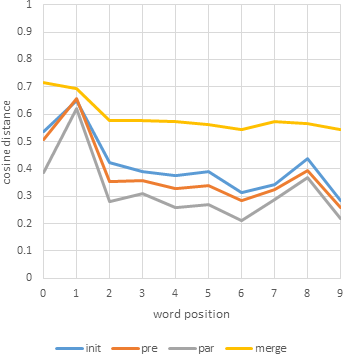}
	}
	
	\subfloat[
		\label{fig:results_convergence_output_min}
		Omission scores: softmax layer (cosine distance).
	]{
		\includegraphics[scale=0.5]{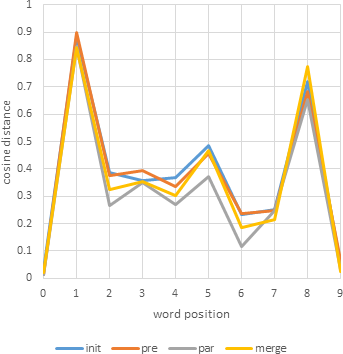}
	}
	\quad
	\subfloat[
	    \label{fig:results_convergence_outputjsd_min}
	       Omission scores: softmax layer (Jensen-Shannon divergence).
	]{
	    \includegraphics[scale=0.5]{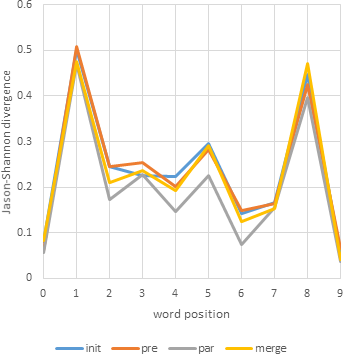}
    }
	\caption{
        \label{fig:results_convergence}
        \small Results for omission scoring of all 9-word long captions (plus the END token). Note that the previous word for position 0 is the START token and position 9 is the END token.
	}
\end{figure}

Although nouns are more frequent at position 8 compared to 1,
there is less sensitivity to the image, across all architectures. This happens even in the merge architecture, which doesn't 
include image features in the RNN hidden state. Hence, this downward trend in gradient is likely due to the caption prefix becoming more 
predictive as it gets longer, progressively reducing the image importance.
Although more sensitive than par-inject, init-inject has a much steeper decline than merge and pre-inject, suggesting that something else is at play apart from prefix information content. 
%
One possibility is that image information is being lost by the RNN as
the prefix gets longer.
To investigate this, we can look at the results for the omission scores which are shown in Fig.~\ref{fig:results_convergence}.

Again, we see peaks at word positions predominately occupied by `visual' words (nouns). The multimodal vector of the merge architecture seems to be an exception. This is because merge's multimodal vector concatenates separate image and prefix vectors, meaning that the image representation is unaffected by the prefix.
The other architectures mix the image and prefix together in the RNN's hidden state vector, requiring the RNN to use memory for both image and prefix, thereby causing visual information to be degraded. 
Note that greater distance between multimodal vectors is unexpected in merge: since image features are concatenated with the RNN hidden state, the RNN part of the multimodal vector is identical in both foil and correct multimodal vectors, which should make the two vectors more similar, not less.

The softmax on the other hand changes very similarly for all architectures and this is reflected both in cosine distance and Jensen-Shannon divergence (compare Fig.~\ref{fig:results_convergence_output_min} and \ref{fig:results_convergence_outputjsd_min}). Merge is slightly more influenced by the image when determining the last noun in the caption and par-inject being the least influenced throughout. The fact that the merge architecture has a very different multimodal vector distance from the other architectures but then ends up with a similar output distance merits further investigation.

\begin{figure}
	\centering
	\subfloat[
		\label{fig:results_stats_convergence_logits_min}
		\small Omission scores: logits (cosine distance).
	]{
		\includegraphics[scale=0.5]{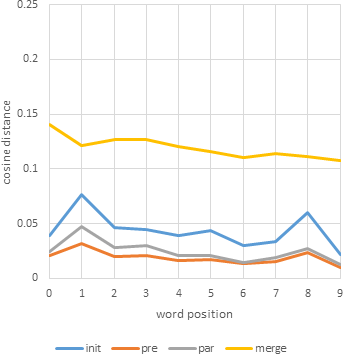}
	}
	\quad
	\subfloat[
		\label{fig:results_stats_numneglogits}
		\small Number of negative numbers in the logits vector.
	]{
		\includegraphics[scale=0.5]{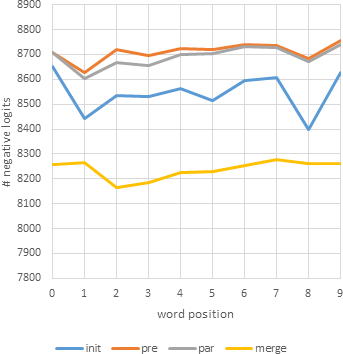}
	}
	\quad
	\caption{
        \label{fig:results_stats}
        \small Further analysis of logits (output layer without softmax) of all 9-word long captions (plus the END token). Note that the previous word for position 0 is the START token and position 9 is the END token.
	}
\end{figure}

We investigate the discrepancy in the results for the merge architecture -- a relatively flat curve for the multimodal vector versus peaks at the output layer -- by
repeating the analysis for the logits vector, that is, the output layer prior to applying the softmax activation function. As shown in Fig.~\ref{fig:results_stats_convergence_logits_min}, this results in curves that are similar to those shown in Fig. \ref{fig:results_convergence_multimodal_min}.

The logits vectors resulting from the original and foil images are much more similar to each other than the multimodal vector for all architectures (peaking at around 0.15 instead of 0.7), but the merge architecture still evinces higher distance between test and foil conditions, and greater stability. The fact that logits are similarly affected by the test-foil discrepancy as the multimodal vectors (compare Fig. \ref{fig:results_convergence_multimodal_min} and \ref{fig:results_stats_convergence_logits_min}) suggests that the peaks observed at the output layer (Fig. \ref{fig:results_convergence_output_min} and \ref{fig:results_convergence_outputjsd_min}) arise from the softmax function itself.
One drastic change that the softmax function performs on the logits vector is the replacement of negative numbers with very small positive numbers. In fact we have found that merge uses fewer negative numbers in the logits vector than other architectures (about 94\% rather than 97\%\footnote{Most logits will be negative since most words will have small probabilities in the softmax.}) as is shown in Figure~\ref{fig:results_stats_numneglogits}, which means that the extra cosine distance between the logits vectors resulting from the original and foil images is probably due to a larger number of elements with opposite signs which, after softmax is applied, become positive and hence more similar. This, coupled with the fact that the output probabilities of any trained caption generator should be similar (otherwise they would not be describing the same image similarly) gives at least a partial explanation for why the outputs in all architectures change similarly when a test image is replaced with the same foil image.

\section{Conclusion}
Caption generators use visual information less and less as the caption is generated, although the amount of visual sensitivity is highly dependent on the part of speech of the word to generate and on the length of the prefix generated so far. This has two implications. 

First, as a caption gets longer, linguistic context becomes increasingly predictive of the next word in the caption, and the image is less crucial. An additional factor, in the case of inject architectures, is that the RNN hidden state stores both visual and linguistic features, making it harder to remember visual features as captions get longer. The evidence for this is that the multimodal vector and logits of the merge architecture change more when the original image to a caption is replaced with a different image, compared to inject architectures.

Second, the peaks observed with nouns in the sensitivity analysis show that image features as currently used in standard captioning models are highly tuned to objects, but far less so to relational predicates such as prepositions or verbs.

For future work we would like to attempt to extract visual words from captions based on how sensitive to the image a trained caption generator is at different word positions in the caption. We would also like to use these techniques to analyze state of the art caption generators, including those with attention, in an effort to deepen our explanation of what makes a good caption generator.

\section*{Acknowledgments}
{\small The research in this paper is partially funded by the Endeavour Scholarship Scheme (Malta). Scholarships are part-financed by the European Union - European Social Fund (ESF) - Operational Programme II – Cohesion Policy 2014-2020 “Investing in human capital to create more opportunities and promote the well-being of society”.}

\clearpage

\bibliographystyle{plain}
\bibliography{bibliography}
\end{document}